\documentclass{article}
\usepackage{spconf,amsmath,graphicx,hyperref}
\usepackage{pifont,xcolor,comment,booktabs}
\usepackage{xspace}

\usepackage{tabularray}
\UseTblrLibrary{booktabs}

\definecolor{lightgrey}{rgb}{0.925, 0.925, 0.925}

\title{KAME: TANDEM ARCHITECTURE FOR ENHANCING KNOWLEDGE \\IN REAL-TIME SPEECH-TO-SPEECH CONVERSATIONAL AI}

\name{So Kuroki, Yotaro Kubo, Takuya Akiba, Yujin Tang}
\address{Sakana AI, Tokyo, Japan.\\
\texttt{\{sokuroki,yotarokubo,takiba,yujintang\}@sakana.ai}
}

\begin{document}
\ninept
\maketitle
\begin{abstract}
Real-time speech-to-speech (S2S) models excel at generating natural, low-latency conversational responses but often lack deep knowledge and semantic understanding. Conversely, cascaded systems combining automatic speech recognition, a text-based Large Language Model (LLM), and text-to-speech synthesis offer superior knowledge representation at the cost of high latency, which disrupts the flow of natural interaction. This paper introduces a novel hybrid architecture that bridges the gap between these two paradigms. Our framework processes user speech through an S2S transformer for immediate responsiveness while concurrently relaying the query to a powerful back-end LLM. The LLM's text-based response is then injected in real time to guide the S2S model's speech generation, effectively infusing its output with rich knowledge without the full latency penalty of a cascaded system. We evaluated our method using a speech-synthesized variant of the MT-Bench benchmark that consists of multi-turn question-answering sessions. The results demonstrate that our system substantially outperforms a baseline S2S model in response correctness, approaching that of a cascaded system, while maintaining a latency on par with the baseline.
\end{abstract}
\section{Introduction}
\label{sec:intro}

Full-duplex conversational speech models are the most promising models for realizing natural interaction between humans and digital assistants \cite{nguyen2023generative,kyutai2024moshi}.
Their ability to continuously listen and respond to user queries opens new possibilities for human-machine interfaces in multiple applications, especially for those that require fast turnaround between humans and machines, such as question answering and brainstorming.

Thanks to the recent developments of large transformers, these models can be implemented as direct speech-to-speech (S2S) models~\cite{fang2024llama, xie2024mini,zhang-etal-2023-speechgpt}.
Because these S2S models are realized by a monolithic architecture that does not need to synchronize with other systems, their turnaround time is typically very low, contributing to more natural interaction.
Moshi \cite{kyutai2024moshi} is a pioneering model in that direction with an end-to-end S2S model for full-duplex conversational AI.

However, because the inputs and outputs of S2S models are information-rich acoustic signals, auto-regressive modeling presents a fundamental challenge related to model capacity.
In other words, unlike in text-only large language models (LLMs), S2S modeling must capture not only the verbal content of speech but also paralinguistic features, such as speaking style, emotion, etc.
This creates a fundamental inefficiency for S2S models with regard to knowledge acquisition.
Given a similar model size, a text-only LLM can learn more knowledge because its capacity is dedicated solely to text, where information is densely represented. An S2S model, by contrast, must expend significant resources on capturing expensive, non-verbal features.
Simply scaling the model is not a straightforward solution, as this presents its own challenges in training stability and resource requirements, particularly because an S2S model inference must operate in real time.

On the other hand, a cascaded architecture excels at knowledge integration.
This approach first transcribes a user's complete utterance using an Automatic Speech Recognition (ASR) model~\cite{zeghidour2025streaming, zusag2024crisperwhisper}.
The resulting text is then fed into a text-based LLM, and the response is synthesized back into speech with a Text-to-Speech (TTS) engine~\cite{zeghidour2025streaming, chatterboxtts2025}.
The primary advantage of this modular design is knowledge extensibility.
That is, the most advanced LLMs can be easily ``plugged in'' to keep the system state-of-the-art.
However, this sequential process introduces latency, a critical drawback in conversational AI.
Because the system must wait for the endpoint of the users' utterances \cite{udupa2025streaming} before the ASR and LLM can even begin their work, the resulting delay disrupts the natural flow of conversation.

\begin{figure}[!t]
  \centering
  \includegraphics[width=1.0\linewidth]{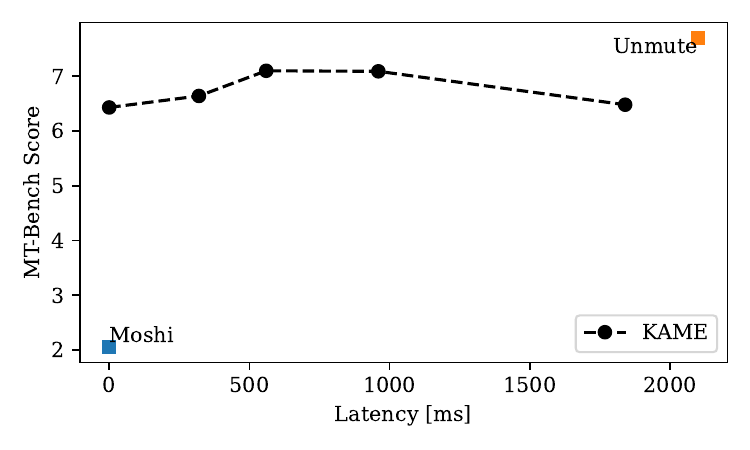}
  \vspace{-0.3 in}
\caption{Performance on MT-Bench vs.\ latency. KAME bridges the gap between low-latency, end-to-end S2S models (e.g., Moshi) and high-quality, cascaded systems (e.g., Unmute).
}
\label{fig:latency_plot}
\vspace{-0.2 in}
\end{figure}

\begin{figure*}[!t]
  \centering
  \centerline{\includegraphics[width=.8\linewidth]{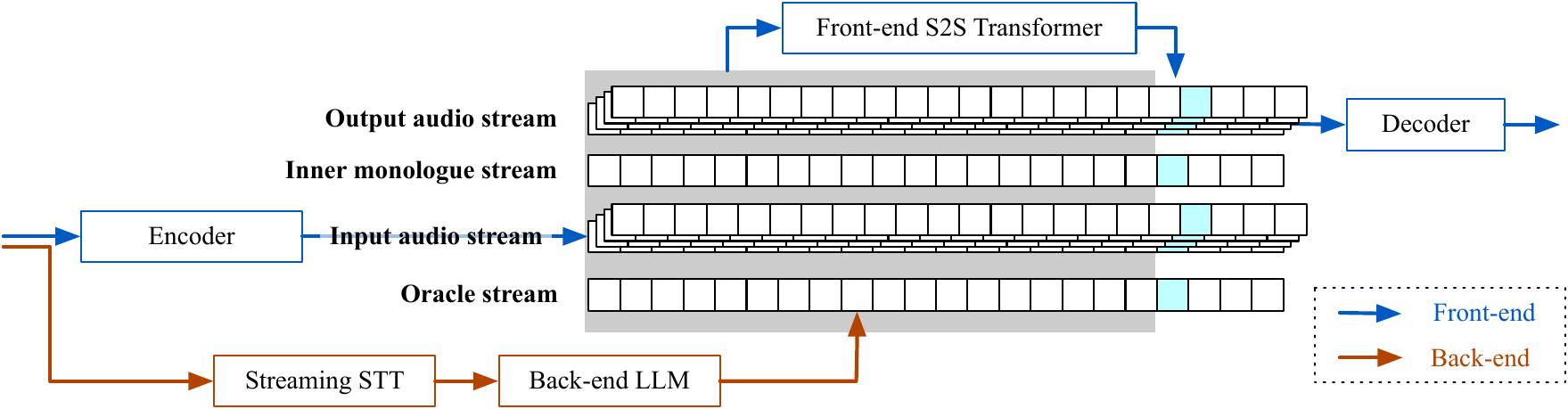}} 
\vspace{-1 mm}
\caption{Proposed architecture for next-token generation.}
\label{fig:res}
\vspace{-0.15 in}
\end{figure*}

To bridge the gap between these paradigms, we introduce a novel hybrid architecture that achieves both the low-latency interaction of S2S models and the knowledge extensibility of cascaded systems.
We call our architecture the \underline{K}nowledge-\underline{A}ccess \underline{M}odel \underline{E}xtension (KAME).
KAME operates as a ``tandem'' system with a front-end S2S model and a back-end text LLM.
The front-end model processes the user's speech in real time for an immediate response.
Simultaneously, it streams an interim transcription to the back-end LLM.
As the LLM formulates a more knowledgeable response, its text is sent back to the front-end through an ``oracle'' stream.
The front-end model is specifically trained to condition its speech output on both its own internal context and this incoming oracle guidance, effectively infusing real-time responsiveness with deep knowledge.
Evaluations on a speech-synthesized variant of the multi-turn MT-Bench~\cite{zheng2023judging} benchmark confirm our approach's effectiveness, showing that KAME achieves conversational quality comparable to the latest cascaded system while preserving the responsiveness of a direct S2S model (see Fig.~\ref{fig:latency_plot}).

Our work is conceptually related to prior research on layered and multi-component architectures for response generation.
Qwen2.5-Omni \cite{xu2025qwen2}, for example, employs a Thinker-Talker architecture that enables real-time synthesis of the text generated by an LLM module (called ``Thinker'' in the paper).
While a layered architecture is advantageous in terms of response time, this still requires complete queries represented in the text.
Their tight integration between ``Thinker'' and ``Talker'' also makes it difficult to switch the back-end ``Thinker'' alone.
Minions \cite{narayan2025minions} realized a cost-efficient response generation in a large-context scenario by splitting the context-handling tasks between a large back-end LLM and local small LMs.
For ASR, the deliberation model \cite{hu2020deliberation} employs a heterogeneous architecture with two components aiming at correcting the results of a smaller front-end model by using a larger back-end system.
Our proposed approach follows this same collaborative strategy.
Its key distinction, however, is that the front-end and back-end components operate on different representations. Concretely, our front-end is an S2S model, while the back-end is a text-based language model.
This loose coupling in our proposed architecture is advantageous because KAME can be made back-end agnostic.
Since frontier LLMs are known to differ widely in their areas of expertise \cite{chang2024survey}, back-end agnosticism enables flexible selection of the back-end LLM and is essential to achieve optimal results in diverse application areas.

In the rest of this paper, we first present our tandem architecture, which enables real-time communication between the front- and back-end components.
Then, we describe the detailed training method for our system.
In the experimental section, we evaluate KAME in MT-Bench and discuss the results.

\section{Tandem Architecture}
\label{sec:architecture}

Fig.~\ref{fig:res} illustrates the architecture of KAME.
Our design consists of two modules that operate on distinct time scales: the front-end S2S module runs at the cycle of discrete audio tokens (e.g., 80 ms), while the back-end LLM is updated at a relatively slower cycle (e.g., 100 - 500 ms).
The key strength of this architecture is that the front-end S2S can generate responses immediately and can refine them using supervision from the back-end (detailed in the following sections).
Unlike cascaded approaches, our method allows the two modules to run independently and remain asynchronously connected.
This design minimizes initial response latency while improving response quality with the assistance of the back-end LLM.

\subsection{Back-end LLM Module}
\label{sec:backend}

The back-end LLM module consists of a streaming speech-to-text (STT) component and a back-end LLM.
As a user speaks, the streaming STT component continuously transcribes their speech and periodically feeds the partial transcripts to the back-end LLM. 
For each partial transcript it receives, the back-end LLM generates a candidate response and sends it to the front-end module.
This architecture allows the front-end model to generate higher-quality, more informed responses by leveraging the superior knowledge and reasoning of the larger back-end LLM in near real time.
The data flow of the back-end module is illustrated by the red arrows in Fig.~\ref{fig:res}.

Specifically, these candidate responses are fed into the front-end S2S transformer as ``oracle'' tokens to guide its output
\footnote{We use the term ``oracle'' to refer to the supervision provided by the back-end LLM, not to manual annotations.}.
A timing challenge arises because the S2S transformer consumes tokens at fixed intervals (e.g., every 80 ms), while the asynchronous responses from the back-end LLM can vary in length and frequency.
This can cause outputs from consecutive calls to overlap.
To resolve this, we prioritize the most recent response, as it is generated from a longer, more complete user transcript and is therefore expected to be the most informative.

\begin{figure*}[!t]
  \centering
  \includegraphics[width=1.0\linewidth]{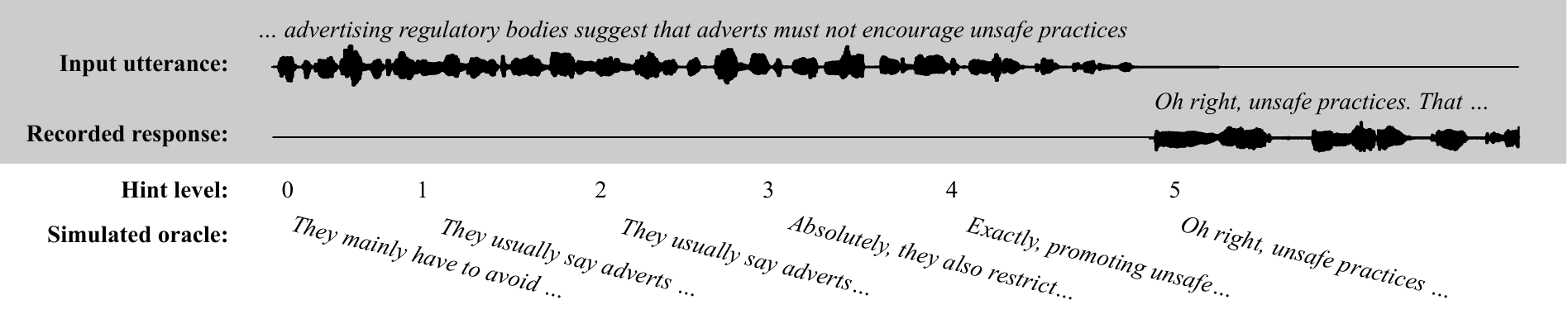}
  \vspace{-20pt}
\caption{An example of generating simulated oracle texts from a recorded conversation. As more of the user's input utterance (top) is revealed over time, the simulated oracle text (bottom) becomes progressively more accurate, and eventually converges to the recorded response (middle). The generation of the oracles is based on the hint level, which in turn depends on the completeness of the partial user speech.}
\label{fig:hint-and-oracle}
\vspace{-0.08 in}
\end{figure*}

\subsection{Front-end S2S Module}
\label{sec:s2s}

Our front-end S2S module adapts the Moshi architecture \cite{kyutai2024moshi}, which consists of three main components: an encoder, an S2S transformer, and a decoder.
The encoder discretizes the continuous audio signal into tokens, which are processed by the S2S transformer.
The decoder then converts the transformer's output tokens back into an audio signal.
The S2S transformer's role is to auto-regressively model multiple interdependent token sequences, or ``streams''.

Building on the three streams in the original Moshi architecture (output audio, inner monologue, and input audio), our KAME architecture introduces a fourth oracle stream.
This new stream incorporates the gradually evolving candidate responses (i.e., oracle tokens) from the back-end LLM, providing guidance to the generation process.
The complete data flow of the front-end module is depicted by the blue arrows in Fig.~\ref{fig:res}.

In this four-stream system, the input audio stream contains the user's discretized speech.
From this, the model generates both the text representation of its speech in the inner monologue stream, and the corresponding discretized audio in the output audio stream.
Our oracle stream introduces external knowledge from the back-end LLM directly into this process.
To effectively integrate this information and overcome timing challenges, the front-end module must be trained using those oracle tokens. We detail a practical method in the following section.

\section{Training Method}
\label{sec:training}

Training a conventional S2S model is straightforward, as any two-party dialogue dataset can be used by treating one user's speech as the input and the other's as the target response.
However, this simplicity does not apply for KAME.
Our architecture requires oracle tokens that not only provide external information but also evolve in real time based on partial user speech.
Since collecting natural conversation data with these specific properties is a significant challenge,
we propose a method to create it by converting a standard two-party dialogue dataset into the required format.

\subsection{Simulated Oracle Augmentation}
\label{sec:simulation}

Since the oracle text doesn't naturally exist, our solution is to simulate it using a standard conversational dataset that already contains a user's input and a corresponding ground-truth response.
The design principle is to make the simulated oracle mimic how a real-time LLM would behave, for example, its predictions should get progressively better as it learns more of the user's speech.
As illustrated in Fig.~\ref{fig:hint-and-oracle}, the process works as follows:
\begin{itemize}
\setlength{\itemsep}{0cm}
    \item \textbf{Early Input:} At the beginning of the user's utterance, the simulated oracle text is a general, plausible sentence. This simulates an LLM that has only heard the first few words and is making an educated guess.
    \item \textbf{Progressive Refinement:} As more of the input utterance is processed, we generate new simulated oracle text that is more specific and closer to the final, correct answer.
    \item \textbf{Final Convergence:} By the time the user finishes speaking, the simulated oracle text is designed to converge and become nearly identical to the actual ground-truth response recorded in the dataset.
\end{itemize}

To generate the simulated oracle text, we use a separate ``simulator'' LLM.
The goal is to control how closely the simulator's output matches the ground-truth response based on how much of the user's input has been heard.
We first quantify the completeness of the user's input at any given timestep $t$ by calculating the ratio of words heard so far and denote it as $r(t) = n(t) / N$, where $n(t)$ is the number of words observed before time $t$ and $N$ is the total number of words in the full utterance.
Based on this ratio $r(t)$, we determine a hint level $\ell(t)$ from 0 to 5, as defined in Table~\ref{table:hints}.
This hint level indicates how we prompt the simulator LLM to generate simulated oracle texts:
\begin{itemize}
\setlength{\itemsep}{0cm}
    \item \textbf{Level 0 ($r(t) \in [0, 0.5)$):} When very little has been heard, the LLM is prompted with only the speech history. It receives no hint and must generate a plausible response from scratch.
    \item \textbf{Level 1-4 ($r(t) \in [0.5, 1)$):} As the input progresses, the LLM receives both the history and the ground-truth response as ``hint'', a specific instruction to guide its usage of the history and refine the simulated oracle sentences.
    \item \textbf{Level 5 ($r(t)=1$):} When the input is complete, no LLM generation is performed. The ground-truth ``hint'' text is used directly as the final simulated oracle sentence.
\end{itemize}

This hint-level mechanism allows us to generate simulated oracle text that starts as an unguided prediction and smoothly converges to the correct response by gradually relaxing the constraints on how it uses the ground-truth hint.
Moreover, by employing a TTS system, text-based question answering datasets can also be used to train our KAME system.
In our experiments, we collected questions and answers on diverse topics and converted them into a conversational style.
We then applied a TTS system and word-aligner to transform the dataset into a form suitable for the procedure above.

\begin{table*}[tb]
\vspace{-6pt}
\caption{Parameters and instructions used for simulated oracle generation. A simulated oracle sentence after obtaining the $n$-th word of an $N$-word utterance is generated with the instruction defined in the $\ell$-th row, where the completeness is defined to be $r(t) = n/N$.}
\label{table:hints}
\centering
\begin{tabular}{ccccl}
\toprule
\multicolumn{2}{l}{\textbf{Progress indicator}} & \multicolumn{3}{l}{\textbf{Information provided for generation (as a prompt)}} \\
Lv. ($\ell$) & Completeness ($r(t)$) & History & Hint & Instruction on hint-usage \\
\midrule
0 & $[0, 0.5)$ & \ding{51} & -- &
N/A \\
1 & $[0.5, 0.65)$ &\ding{51} &   \ding{51} &
\texttt{Refer only to keywords from the hint string.} \\
2 & $[0.65, 0.8)$ &\ding{51} &  \ding{51} &
\texttt{Include content different from the hint.} \\
3 & $[0.8, 0.95)$ &\ding{51} &   \ding{51} &
\texttt{Don't copy the hint verbatim. }  \\
4 & $[0.95, 1)$ &\ding{51} &  \ding{51} &
 \texttt{Use the hint.} \\
5 & $[1, 1]$ & -- &  -- & 
 N/A; No LLM generation performed. ``Hint'' is directly used as the generated text. \\
\bottomrule
\multicolumn{5}{r}{{\scriptsize  History: All previous words up to timestamp $t$ in the session. Hint: Transcription of the next response.}} \\
\end{tabular}
\vspace{-0.3 in}
\end{table*}

\subsection{Training Front-end S2S Model}

With the data generation complete, we can now train the front-end S2S module.
The training data consists of audio inputs paired with the corresponding simulated oracle sentences, which are injected at fixed intervals to mimic their real-time arrival.
Several key modifications were made to prepare this data for training:
(1) We tokenize the simulated oracle sentences using the same tokenizer as the inner monologue stream, ensuring consistency within the model;
(2) To help the model distinguish between consecutively arriving simulated oracle sentences, each one is prefixed with a dedicated special token that marks its boundary;
(3) To improve robustness, we add random jitter to the arrival timing of the simulated oracle tokens during training, simulating the small delays and variations that would occur in a live system.
For the training objective, we use a combined loss function of text and audio, weighting the audio loss by a factor of 1.5.
All other hyper-parameters and settings are identical to the original Moshi paper.

\section{Experiments}
\label{sec:experiments}

We evaluate KAME and compare it with several baselines.
Our baselines include Moshi \cite{kyutai_moshi_pretrained}, a representative S2S system, and Unmute  \cite{kyutai_unmute}, a pioneering cascaded system.
We modified Unmute's back-end so that it works with external LLMs for the sake of comparisons.

\subsection{Experimental Setups}

To construct the base conversational dataset, we seed from benchmark Q\&A pairs and generate synthetic dialogue sessions: 22,800 from 11,996 MMLU-Pro entries~\cite{wang2024mmlu}, 11,742 from 7,428 GSM8K entries~\cite{cobbe2021training}, and 22,040 from 12,012 HSSBench entries~\cite{kang2025hssbench}.
We then add simulated oracles to these datasets. We fine-tuned Moshi on the same base conversational data for fair comparison.

We compared two metrics across these systems: Response latency and answer quality over a subset of the MT-Bench benchmark \cite{zheng2023judging}.
We selected the subset by excluding the question categories that are not suitable for speech interaction.
Specifically, ``Coding'', ``Extraction'', ``Math'', ``Roleplay'', and ``Writing'' are excluded from the evaluation.
Response latency is computed by measuring the duration between the end of a user's utterance (question) and the beginning of a system's utterance (answer).
Following the standard procedure described in the MT-Bench paper, we adopt ``LLM-as-a-Judge'' to measure the quality of the answers.
All experiments are conducted six times, and we report the average results.

\begin{table}[tb]
\vspace{-6pt}
\caption{MT-Bench scores. Scores over reasoning (reas.), STEM (stem), humanities (human.) subsets of MT-Bench. Turn one and two scores are averaged. The ``avg.'' column contains the averages of all scores of the subsets extracted. ``latency'' is the median response latency in seconds.}
\label{table:score_and_latency}
\centering
\begin{booktabs}{lccccc}
\toprule
Method & reas. & stem & human. & avg. & latency \\
\midrule
Moshi           & 1.32 & 1.94 & 2.88 & 2.05 & 0.0 \\
KAME (Ours)            & \\
\xspace w/ GPT-4.1             & 5.44 & {\bf 6.65} & {\bf 7.19} & {\bf 6.43} & 0.0 \\
\xspace w/ Claude-opus-4.1     & {\bf 5.72} & 6.53 & 6.43 & 6.23 & 0.0 \\
\midrule[dashed] Unmute (w/ GPT-4.1)         & 7.08 & 8.35 & 7.67 & 7.70 & 2.1 \\
\bottomrule
\end{booktabs}
\vspace{-10pt}
\end{table}

\subsection{Results and Discussions}

Table \ref{table:score_and_latency} summarizes the MT-Bench scores and latencies in our experiments.
We observe that our proposed architecture (KAME) significantly outperforms Moshi in question answering.
The MT-Bench score improved from 2.05 (Moshi) to 6.43 (KAME) by adding external LLMs as a knowledge-access model.
The median latency was confirmed to be the same after introducing the knowledge-access model.
In other words, for both Moshi and KAME, for at least half of the sessions, the model starts responding before the questions end.

KAME also proves to be back-end agnostic.
Our training was done using GPT-4.1-nano as a primary back-end, however, the KAME's two rows indicate that swapping the back-end models does not hurt the performance significantly.
We also observe that the category-wise relative advantages differ depending on the back-end LLM choice.
With KAME, one can choose the optimal back-end model depending on the context in which it would be deployed.

In comparison with a cascaded system, our method is significantly faster than Unmute.
However, the MT-Bench score is degraded compared to it.
This is mainly because our method starts responding with significantly less information.
Regardless, we notice that KAME can correct its speech content after getting contradictory oracles from the external model.
While this is a natural and desired behavior in real-world conversations, this also produces redundant expressions in written text that affect the LLM-as-a-Judge evaluation scheme, and result in a degradation in the MT-Bench score.

In principle, a cascaded system should be able to leverage the full capability of the back-end LLMs if there are no speech recognition errors.
On the other hand, our system can be degraded due to the premature response generation.
To better understand the capability of the back-end LLMs in isolation, we retrieve sentences of the last oracle injection in each session from the experiments of KAME with GPT-4.1, and evaluate the back-end's text responses directly.
The results are: reasoning 6.48, STEM 8.34, and humanities 8.56, with an average of 7.79.
The cascaded Unmute system (with GPT-4.1) achieves comparable scores, whereas KAME (with GPT-4.1 as the back-end LLM) performs worse.
This implies that the main reason for KAME's gap to cascaded systems is not the back-end LLM's capability but the timing of its early responses.

To analyze the trade-off between response latency and quality, we conducted an experiment where we forced our model to delay its response.
This is achieved by compelling the S2S model to output only silence tokens (for audio) and padding tokens (for the inner monologue) until a specified time has passed.
The results, shown in Fig.~\ref{fig:latency_plot}, plot the MT-Bench scores as a function of this forced delay.
Initially, delaying the response improves the model's quality.
However, the benefit diminishes in higher-latency settings, a limitation we attribute to the lack of long-pause examples in our training.
This could be mitigated by corresponding data expansion and curation in the future.
Most importantly, this analysis verifies that the performance gap to a cascaded model is due to premature generation, confirming that KAME strikes an effective balance between the competing demands of low latency and high-quality output.

\section{Conclusions}

In this paper, we introduced KAME, a novel tandem architecture for real-time conversational AI.
KAME successfully integrates the advanced capabilities of an LLM with the low latency of a direct S2S system via the introduction of oracle tokens.
We also proposed practical training methods for this architecture, including a technique for generating the necessary synthetic oracle data.

Our experiments show that KAME significantly outperforms conventional approaches.
Compared to a cascaded system, KAME dramatically reduces latency while keeping quality degradation minimal.
Conversely, compared to a standard S2S model, our method substantially improves response quality without increasing latency.
Our results demonstrate that this tandem architecture provides an effective and balanced solution for building powerful, real-time conversational AI.
An interesting challenge for future work is to extend KAME to support scenarios where the front-end model alone does not perform well, such as multi-party conversations.

\vfill\pagebreak

\bibliographystyle{IEEEbib}
\bibliography{strings,refs}

\end{document}